%% file: main.tex
\documentclass[sigconf]{acmart}
\AtBeginDocument{%
  }

\setcopyright{acmlicensed}
\copyrightyear{2025}
\acmYear{2025}
\acmDOI{XXXXXXX.XXXXXXX}
\acmConference[CI'2025]{ACM Collective Intelligence}{2025}{CA, USA}
\acmISBN{978-1-4503-XXXX-X/2018/06}

\begin{document}

\title{Measuring Implicit Spatial Coordination in Teams: Effects on Collective Intelligence and Performance}

\author{Thuy Ngoc Nguyen}
\affiliation{%
  \institution{University of Dayton}
  \city{Dayton}
  \state{Ohio}
  \country{USA}
}
\email{ngoc.nguyen@udayton.edu}
\author{Anita Williams Woolley}
\affiliation{%
  \institution{Carnegie Mellon University}
  \city{Pittsburgh}
  \state{Pennsylvania}
  \country{USA}
  \postcode{15213}}
  \email{awoolley@andrew.cmu.edu}

\author{Cleotilde Gonzalez}
\affiliation{%
  \institution{Carnegie Mellon University}
  \city{Pittsburgh}
  \state{Pennsylvania}
  \country{USA}
  \postcode{15206}}
  \email{coty@cmu.edu}
  

\renewcommand{\shortauthors}{Nguyen et al.}

\begin{abstract}
Coordinated teamwork is essential in fast-paced decision-making environments that require dynamic adaptation, often without an opportunity for explicit communication. Although implicit coordination has been extensively considered conceptually in the existing literature, the majority of work has focused on co-located, synchronous teamwork (such as in sports teams) or, in distributed teams the focus has been primarily on coordination of knowledge work. However, many teams (firefighters, military, law enforcement, emergency response) need to coordinate their movements in physical space without the benefit of visual cues or extensive explicit communication.  This paper investigates how three dimensions of spatial coordination, namely exploration diversity, movement specialization, and adaptive spatial proximity, influence team performance in a collaborative online search and rescue task where explicit communication is restricted, and team members must rely on movement patterns to infer others' intentions and coordinate their actions. Our metrics capture the relational aspects of teamwork by measuring spatial proximity, distribution patterns, and alignment of movements within shared environments. We analyze data from an experiment including 34 four-person teams (136 participants) assigned to specialized roles in a search and rescue task. Our results demonstrate that spatial specialization positively predicts team performance, while adaptive spatial proximity exhibits a marginal inverted U-shaped relationship, suggesting that moderate levels of adaptation are optimal. Furthermore, the temporal dynamics of these spatial coordination metrics clearly differentiate high- from low-performing teams over time. These findings provide insights into implicit spatial coordination in role-based teamwork and highlight the importance of balanced adaptive strategies, with implications for team training and the development of AI-assisted team support systems.

\end{abstract}

\ccsdesc[500]{Human-centered computing~Collaborative and social computing theory, concepts and paradigms}
\ccsdesc[300]{Computing methodologies~Simulation evaluation}
\ccsdesc[300]{Computing methodologies~Cognitive science}

\keywords{Implicit Coordination, Collective Intelligence, Spatial Behavior, Process Metrics, Adaptive Coordination, Search and Rescue Simulation}

\maketitle

\section{Introduction}
\label{sec:intro}
\input{1.intro}

\section{Related Work}
\label{sec:related_work}
\input{2.related-work}

\section{Method}
\label{sec:method}
\input{3.method}

\section{Analysis}
\label{sec:results}
\input{4.results}

\section{Discussion}
\label{sec:discussion}
\input{5.discussion}

\begin{acks}
This research was supported by the Defense Advanced Research Projects Agency and was accomplished under Grant Number W911NF-20-1-000. Thuy Ngoc Nguyen was supported by the University of Dayton Research Council Seed Grant.
\end{acks}

\bibliographystyle{ACM-Reference-Format}
\bibliography{references.bib}
\appendix
\input{6.appendix}
\end{document}

%% file: 1.intro.tex
Teams have become the fundamental building blocks of modern organizations, increasingly used to address complex and dynamic challenges that require the integration of the knowledge and skills of individual experts to address immediate challenges ~\cite{hackman1987} and facilitate innovative solutions~\cite{wuchty2007increasing,uzzi_atypical_2013}. Collective intelligence (CI) is the general ability of a group to work together effectively on a variety of tasks~\cite{woolley2010evidence}. Although coordination is studied extensively in the extant literature on group and team performance, much of that work is focused on explicit coordination through verbal communication.  By contrast, relatively less research has focused on tacit or implicit coordination, or the synchronization of members' actions based on unspoken assumptions about what others in the group are likely to do~\cite{wittenbaum_tacit_1996}. In particular, in spatially distributed tasks where team members must navigate shared environments, implicit spatial coordination becomes essential, especially when the time or opportunity for explicit communication is limited~\cite{demir2018team}. Existing research has shown that collective intelligence is associated with various forms of synchrony in nonverbal behavior in co-located teams \cite{chikersal2017deep, tomprou_speaking_2021, woolley_collective_2023}, as well as with temporal coordination of communication behavior in distributed teams~\cite{mayo_variance_2021, riedl_teams_2017}. 
However, very little work has focused on ways to capture the coordination of behavior in physical space, or spatial coordination, when members perform the same or compatible actions at the same time in varying spatial configurations \cite{feese_sensing_2014}.  Given how essential such coordination is in situations encountered by military, law enforcement, and emergency response crews, to name a few, it is important to examine and validate ways to capture the quality of spatial coordination in order to better understand how to support its development. 

Research on team implicit coordination has significantly advanced our theoretical and methodological understanding of team interactions~\cite{espinosa2004explicit,macmillan2004communication,rico2008team, rico_building_2019}. Existing studies have identified two essential components of implicit coordination, namely anticipation, where team members predict each other's needs and actions, and dynamic adjustment, where members adapt behaviors accordingly~\cite{rico2008team}. Team situation models (TSMs), emergent team-level knowledge structures, have been proposed as the underlying mechanisms that allow implicit coordination, aligning closely with established findings on shared mental models that facilitate effective team processes~\cite{mathieu2000influence} and cognitive models of situation awareness~\cite{endsley2000theoretical}. However, despite the convergence of the literature around this theoretical framework, there is still a need for measures that effectively capture spatial coordination, or alignment of members' movement in physical space, particularly when it is implicit as a result of limitations on explicit communication. Given extant research demonstrating a relationship between CI and other forms of physical synchrony \cite{chikersal2017deep, tomprou_speaking_2021, woolley_collective_2023} we would anticipate a relationship with implicit coordination in spatial contexts as well, however, it is important to examine what nuances may exist to gain a better understanding of how this capability develops in collective coordination.


In this work, we investigate how spatial coordination influences team performance in settings where explicit communication is restricted. Building on Rico et al.'s framework of implicit coordination~\cite{rico2008team}, we introduce new spatial coordination metrics that directly capture the relational aspects of teamwork. We focus specifically on role-specialized coordination in spatial navigation tasks, providing quantifiable measures of proximity, distribution patterns, and movement alignment that reflect implicit coordination beyond simple aggregation of individual behaviors.
In particular, our primary research questions are:
\begin{itemize}
    \item \textbf{RQ1}: How do different dimensions of spatial coordination predict CI and team performance?
    \item \textbf{RQ2}: To what extent does CI mediate the relationship between spatial coordination and performance?
    \item \textbf{RQ3}: How do the temporal dynamics of coordination patterns differentiate high- from low-performing teams?
\end{itemize}

To investigate our research questions, we collected data from 136 participants randomly assigned to one of 34 four-person teams who collaborated on an online search and rescue task in a shared two-dimensional online map environment. The task involved two different specialized roles, and the mission involved tasks with different levels of interdependence; some required sequential coordination, where one role needed to act before another could complete a rescue; others required simultaneous coordination, where different roles were required to be co-located for a specific amount of time to complete a rescue; and a third subset of tasks could be completed by team members independently. 

We developed three metrics to capture the quality of spatial coordination: (1) Spatial Exploration Diversity (SED), calculated via Jensen-Shannon divergence between movement probability distributions; (2) Spatial Movement Specialization (SMS), which captures the balance between exploration complexity and effective division of labor; and (3) Spatial Proximity Adaptation (SPA), which tracks changes in inter-role physical distance between task phases. These metrics were designed to capture the diversity in how members explore the space (diversity), their level of differentiation or specialization in searching the space (specialization), and their adjustment in spatial proximity based on task demands (adaptation). 

To analyze these dynamics, we employ a multi-method quantitative approach to examine the direct effect of spatial coordination, based on these metrics,  and its mediated impact through CI using bootstrapped mediation analysis. We further explore curvilinear relationships through nonlinear analyses, investigating whether optimal levels of adaptive spatial proximity exist and how these patterns evolve over time in high- versus low-performing teams.


Our results show that Spatial Movement Specialization (SMS) significantly predicts both CI and overall team performance, while Spatial Exploration Diversity (SED) and Spatial Proximity Adaptation (SPA) did not exhibit significant linear relationships with performance or CI. We also found that CI partially mediated the SMS-performance link, accounting for 47.6\% of the total effect. Interestingly, SPA demonstrated a marginal inverted U-shaped relationship with team performance, suggesting that an optimal level of adaptive proximity exists. Furthermore, temporal dynamics across all three spatial metrics clearly differentiated high- from low-performing teams. These findings highlight how role-based spatial organization and temporal adaptability shape team performance in spatially distributed environments where explicit communication is limited. We discuss the implications of these results for improving team coordination.

%% file: 2.related-work.tex
\paragraph{\textbf{Implicit coordination}} Research on implicit coordination has progressed from identifying core mechanisms to understanding factors that facilitate its emergence across diverse team settings. Foundational work established anticipation and dynamic adjustment as key components of implicit coordination~\cite{rico2008team}, while other studies demonstrated that teams often transition from explicit to implicit coordination under high workload conditions~\cite{entin1999adaptive}. Despite these theoretical advances, a methodological gap remains in how implicit coordination is measured as it unfolds in real time. Prior studies have largely relied on indirect assessments, such as communication analyses~\cite{butchibabu2016implicit}, shared mental model evaluations~\cite{mathieu2000influence}, or outcome-based metrics in distributed team contexts~\cite{espinosa2007team}. While these approaches are valuable, they do not fully capture the dynamic, behavioral manifestations of coordination during task execution. We argue for the need for quantitative spatial metrics that capture real-time coordination of movement and positioning among team members in collaborative environments. 

\paragraph{\textbf{Spatial coordination behavior}} Research across multiple domains has investigated spatial behavior as an indicator of team cognition, revealing how movement patterns reflect underlying cognitive processes.  In military contexts, studies have highlighted the importance of adaptability and coordination in response to changing battlefield conditions~\cite{salas2008wisdom}. Medical teams employ adaptive coordination strategies during surgical procedures, which are associated with shared mental models~\cite{kolbe2013co}. Similarly, sports science research has identified how player positioning and anticipation of movement serve as physical manifestations of team coordination~\cite{bourbousson2010team,eccles2004expert}. These spatial patterns are increasingly seen as observable traces of cognitive alignment that would otherwise remain hidden. Prior research argues that team cognition is inherently ``situated'' and best understood through its behavioral manifestations rather than aggregated individual knowledge~\cite{cooke2013interactive,cooke2024teams}. However, while these studies acknowledge the importance of spatial behavior, they typically employ qualitative coding schemes or focus on static analyses rather than dynamic movement relationships. Our work differs by providing quantitative metrics that directly capture relational aspects of spatial coordination through mathematical formulations of pattern diversity, role-based coordination, and adaptive changes over time. 

\paragraph{\textbf{Collective Intelligence (CI)}} Research on CI has established it as a critical mediating factor linking team processes to performance outcomes. Seminal research demonstrated the existence of a ``c factor'', a general ability of groups to perform effectively across diverse tasks~\cite{woolley2010evidence}. 
Existing research has shown that measures of a team's CI predict future team performance~\cite{engel2015collective, kim2017makes}. Previous studies have primarily focused on communication patterns~\cite{woolley2015collective}, social perceptiveness~\cite{engel2014reading}, and physiological synchrony~\cite{chikersal2017deep} as antecedents of CI. Research has also explored CI as a mediating mechanism connecting the diversity of cognitive styles with implicit team learning~\cite{aggarwal2019impact}. A comprehensive meta-analysis involving over 1300 groups found that collaborative process behaviors are stronger predictors of CI than individual member skills or group composition~\cite{riedl2021quantifying}. Relatedly, interaction quality has been identified to play an important role in the emergence of CI~\cite{janssens2022collective}, and recent research has highlighted the need to understand the dynamic processes by which groups adapt to changing environments~\cite{galesic2023beyond}. Recent research has used a set of individual-level collaborative process metrics to predict collective intelligence (CI) and evaluate collective outcomes in spatial navigation tasks~\cite{zhao_teaching_2023,mcdonald2023working}. However, this may overlook the nuanced coordination strategies and emergent interaction patterns critical to the dynamic evolution of CI. Thus, we explore metrics that directly capture the relational and dynamic aspects of team members' movements and spatial positioning. By quantifying these relational dynamics, we examine how spatial coordination influences team performance and whether this relationship is mediated by CI. This is particularly relevant in contexts where explicit communication is restricted, requiring team members to infer others' intentions and coordinate their actions primarily through movement patterns.

%% file: 3.method.tex
\begin{figure}[!htbp]
\centering
 \includegraphics[width=0.95\linewidth]{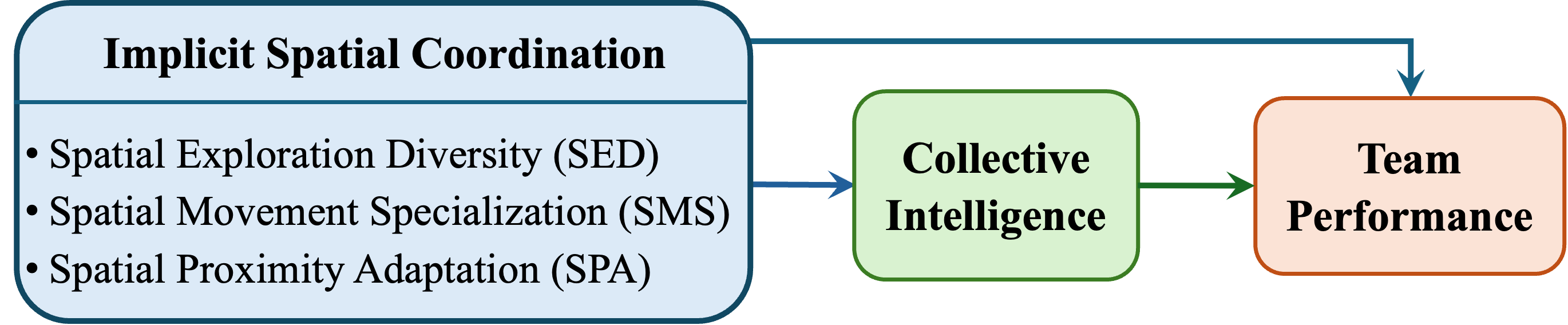}
\caption{General research model.}
\label{fig:overiew-method}
\end{figure}

Our study investigates how implicit coordination in spatial contexts influences team performance, with collective intelligence as a potential mediating mechanism. Building upon prior work defining implicit coordination as anticipatory team interactions without explicit communication~\cite{rico2008team}, we explore three spatial metrics to quantify coordination patterns: spatial exploration diversity (SED), spatial movement specialization (SMS), and spatial proximity adaptation (SPA). These metrics address a gap in understanding how spatial coordination unfolds in environments where explicit communication is limited. In addition, based on research on observable indicators of collective intelligence~\cite{riedl2021quantifying}, we examine whether these spatial coordination metrics would predict both collective intelligence and team performance outcomes. We employ a multi-method quantitative approach to examine both direct effects and mediated pathways through collective intelligence, as depicted in Figure~\ref{fig:overiew-method}. This approach allows us to identify which spatial coordination behaviors contribute to team performance and to understand the mechanisms through which these behaviors operate.

\subsection{Experimental Task}
\begin{figure*}[t]
    \centering
    \begin{minipage}[t]{0.49\textwidth}
        \centering
        \includegraphics[width=\textwidth]{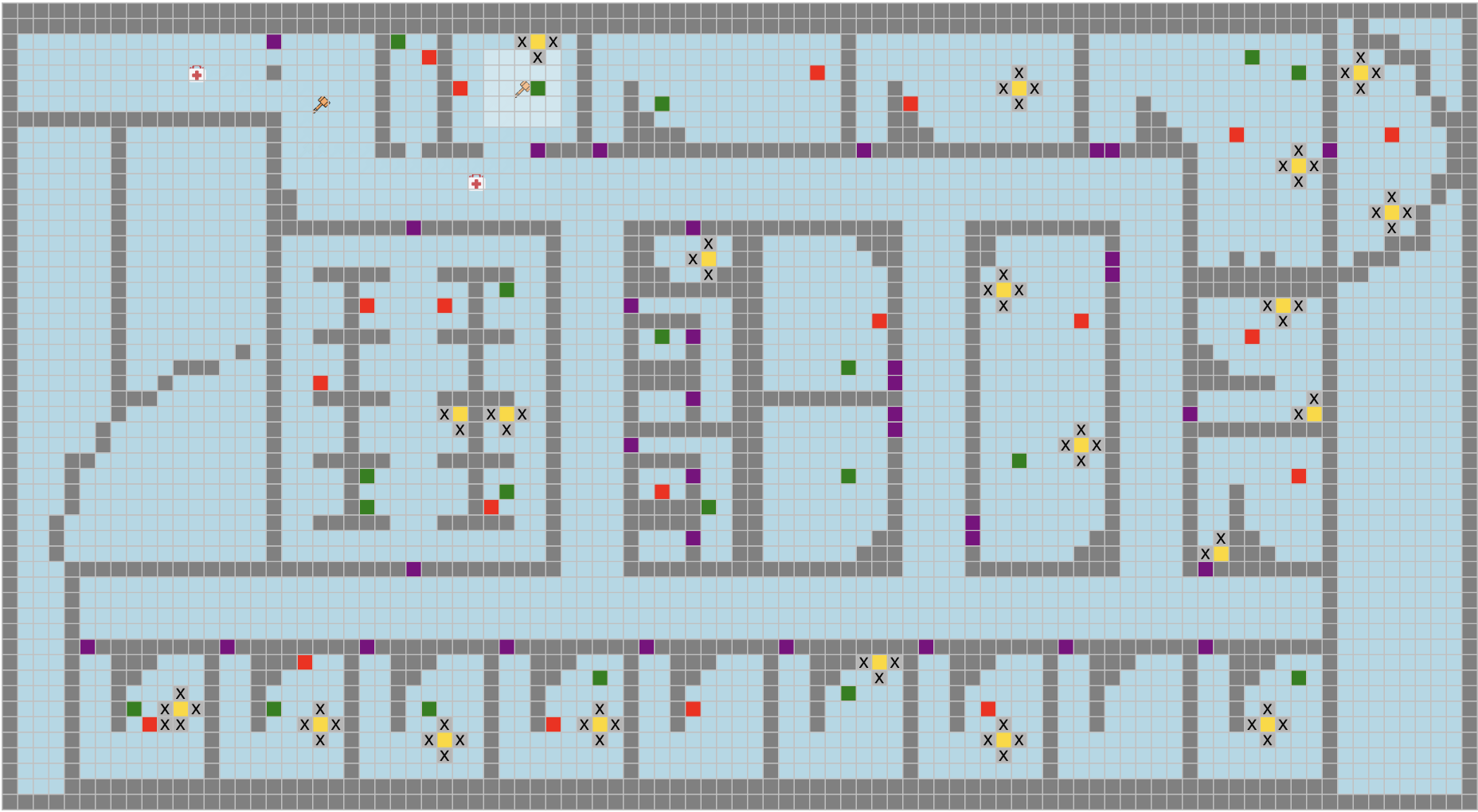}
        \label{fig:fullview}
    \end{minipage}
    \hfill
    \begin{minipage}[t]{0.49\textwidth}
        \centering
        \includegraphics[width=\textwidth]{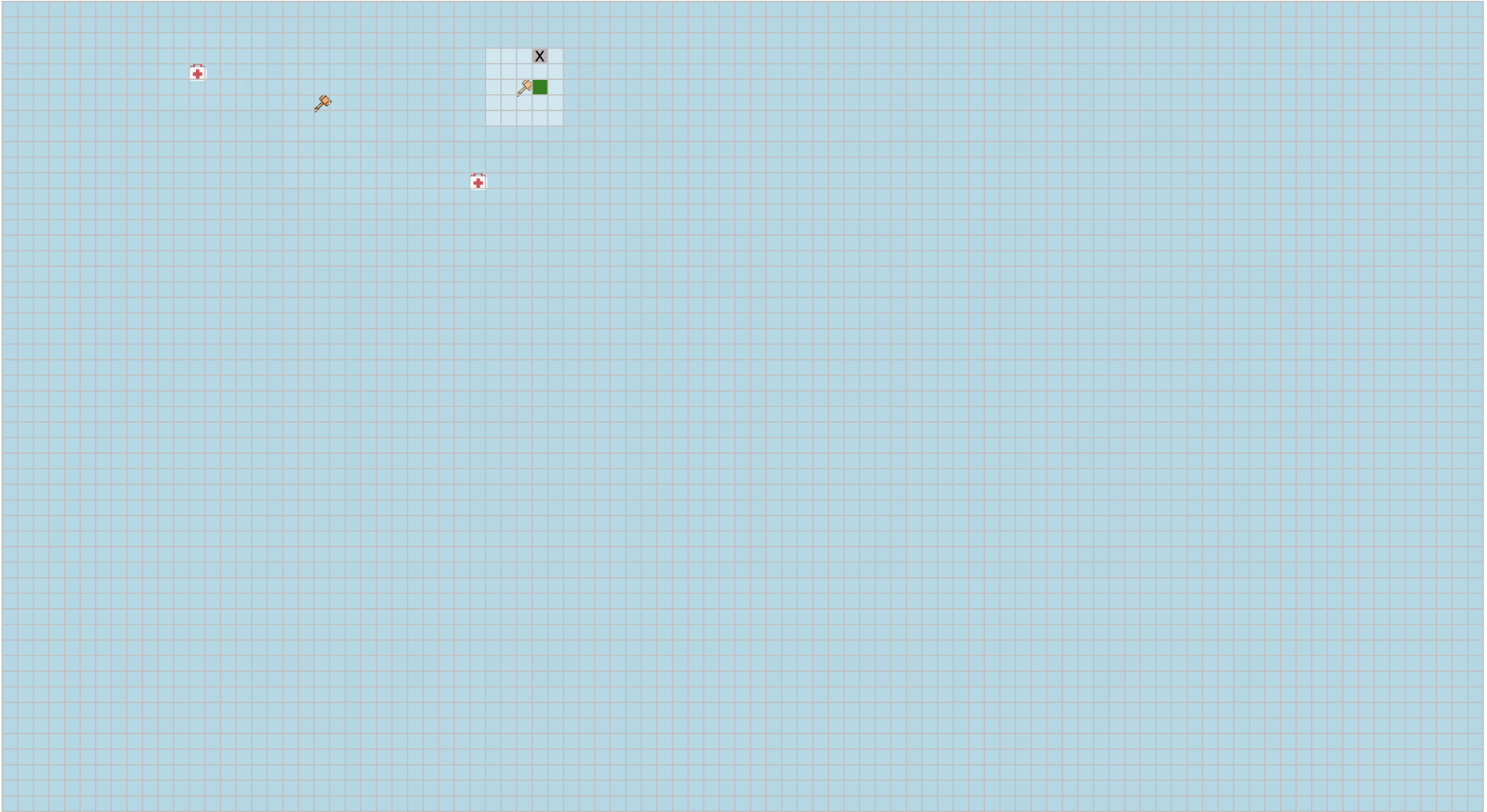}
        \label{fig:fov}
    \end{minipage}
    \vspace{-0.5cm}
    \caption{Team Minimap with the left panel shows the complete map overview with all victims and obstacles visible, while the right panel shows participants' actual restricted field of view (limited to five squares at a time).}
    \label{fig:team-minimap}
\end{figure*}

We use the Team Minimap task, an online, multi-player adaptation of the search and rescue scenario in the Minimap environment~\cite{nguyen2023minimap}, to assess coordination in teams of four members (two medics and two engineers) operating in a complex spatial environment without communication~\cite{mcdonald2023working}. 
Team roles are specialized: medics can rescue green, yellow, and red victims, but rescuing red victims requires an engineer to be adjacent. Green and yellow victims can be rescued at any time during the 5-minute mission, while red victims must be rescued within the first three minutes. Engineers can independently rescue green victims, open doors, and clear obstacles around yellow victims to enable medic access. The team's goal is to maximize accumulated points through implicit coordination. Figure~\ref{fig:team-minimap} illustrates the task environment.

\begin{figure}[!htbp]
\centering
 \includegraphics[width=\linewidth]{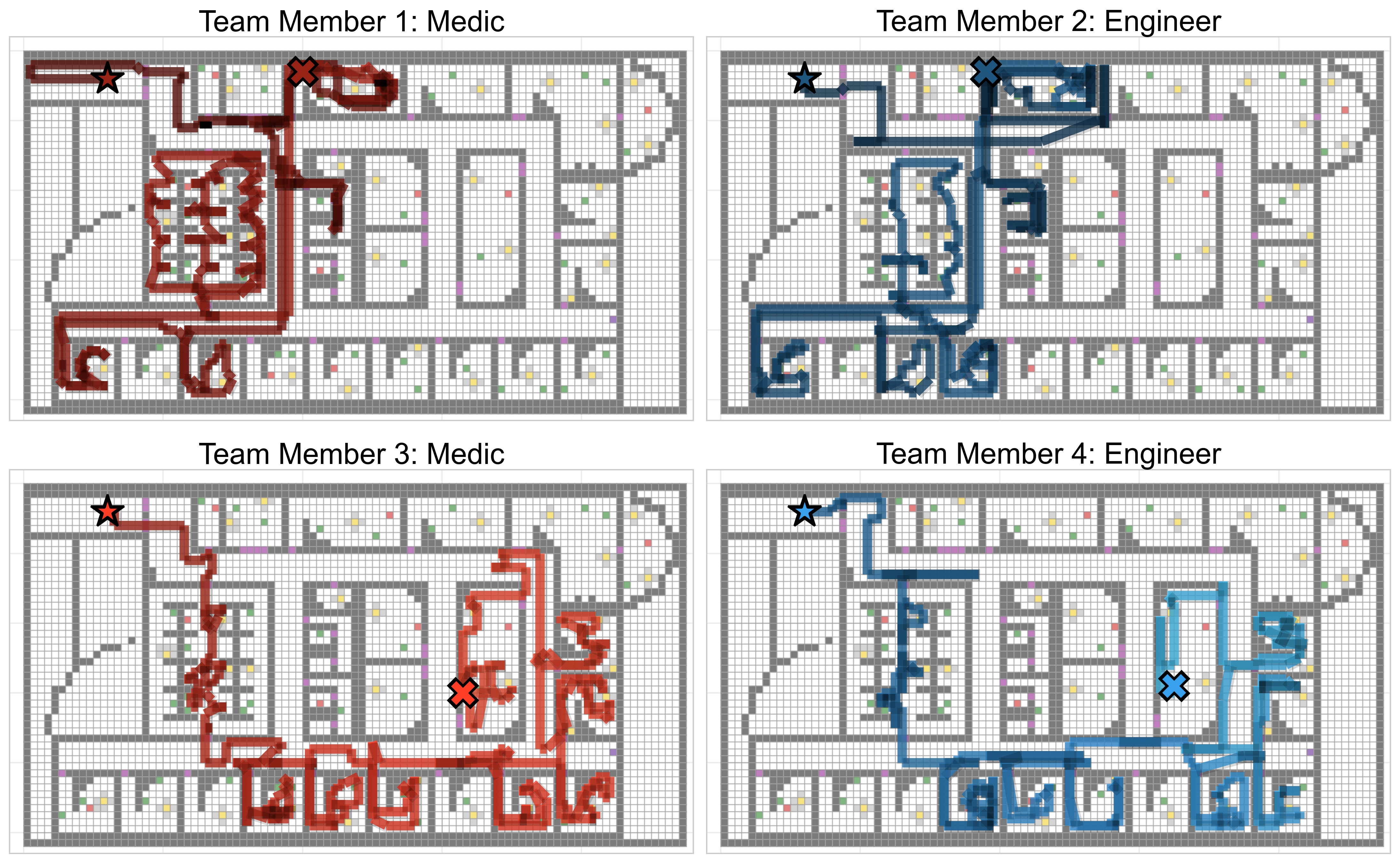}
\caption{Trajectory visualization of a four-member team during the mission, with engineers (blue paths, right side) and medics (red paths, left side). All team members began at a common starting point, marked by stars.}
\label{fig:visualization}
\end{figure}

Figure~\ref{fig:visualization} illustrates the movement trajectories of a four-person team navigating the Team Minimap environment. The paths illustrate how team members distributed themselves across different map sections and converged at some points to perform coordinated rescues. These trajectories reveal how team members explored and coordinated within the task.

\subsection{Study Design and Procedure} 
\subsubsection{Participants}
The sample consists of 136 participants, randomly assigned to 34 teams of four. Participants were recruited from Amazon Mechanical Turk in the United States and compensated for their participation. The mean age was 40.07 years (SD = 11.25), and 53.13\% identified as male. The study was conducted in accordance with ethical guidelines for research involving human subjects and was approved by our institution’s Institutional Review Board. All participants provided informed consent online.
\subsubsection{Design}
All team members played on the same two-dimensional map. Each participant's field of view was limited to a radius of five squares around their avatar, restricting their visibility of the overall environment. Participants could also observe an icon indicating the role and location of each teammate.

The task included three types of victims: green (minor), yellow (serious), and red (critical) to capture different levels of rescue priority. Rescuing green, yellow, and red victims yielded 10, 20, and 30 points, respectively. Participants earned points for their team by rescuing victims, and the total number of points earned across the two rounds determined the team’s bonus payment.

The design includes two forms of coordination to represent different levels of role interdependence. First, sequential coordination is required for yellow victims, which are always trapped by rubble. Engineers need to clear the rubble before medics can complete the rescue. Second, simultaneous coordination is required for red victims, which are only available during the first three minutes of each mission. Rescuing a red victim required both an engineer and a medic to be adjacent to the victim at the same time.

\subsubsection{Procedure}
In the study, participants first provided demographic information and then read the task instructions before being randomly assigned to groups of four. Each group completed two identical 5-minute sessions of the Team Minimap task. The first session served as a practice round to help participants become familiar with the task environment and controls. For all analyses, only data from the second session were used. After completing the task, participants filled out a questionnaire designed to assess their experience and their perceptions of team coordination.

\subsection{Measures}
This section outlines the quantitative metrics used to evaluate implicit spatial coordination, collective intelligence, and team performance during the experimental task. All metrics were computed at 3-second intervals throughout the second of two 5-minute missions, with the first mission serving as a practice session for participants to familiarize themselves with the task.

\subsubsection{Implicit Spatial Coordination Metrics}
To quantify how team members coordinate their movements and activities in space without explicit communication in the search and rescue task, we developed three metrics that capture different aspects of spatial coordination between team members. 
\paragraph{\textbf{Spatial Exploration Diversity (SED)}} SED measures how differently team members explore the environment. This metric uses Jensen-Shannon divergence to calculate dissimilarity between players' movement patterns. We divide the environment into an $m \times n$ grid and represent each player's movements as a probability distribution over grid cells.

For each player, we create a probability distribution $P_i$ where each element represents the frequency of visits to a specific grid cell. The SED is calculated as the average Jensen-Shannon divergence between all pairs of player distributions within the same team:

\begin{equation}
\text{SED} = \frac{1}{N(N-1)/2}\sum_{i=1}^{N-1}\sum_{j=i+1}^{N} \text{JSD}(P_i, P_j),
\end{equation}
where $\text{JSD}(P_i, P_j)$ is the Jensen-Shannon divergence between the movement distributions of players $i$ and $j$, and $N$ is the number of players in a team (typically 4 in our study). The indices $i$
and $j$ represent any two different players within the same team, regardless of their role (engineer or medic). Higher values indicate greater diversity in spatial exploration strategies.

\paragraph{\textbf{Spatial Movement Specialization (SMS)}} 
Inspired by prior research on collaborative tasks and coordination mechanisms~\cite{entin1999adaptive,riedl2021quantifying}, effective team coordination in spatial tasks requires balancing two factors: equitable workload distribution where team members maintain similar levels of engagement and thoroughness, and (2) efficient spatial division of labor to minimize redundant coverage. To capture this balance, the Spatial Movement Specialization (SMS) metric quantifies how effectively specialized roles coordinate their movements in space. It was formulated as the product of two components that reflect these aspects, as follows:


\begin{equation}
\text{SMS} = E_s \times (1 - O).
\end{equation}
Here, $E_s$ is the entropy similarity between role distributions and $O$ is the spatial overlap between roles.
For each role type $r$, we first aggregate movement data across all players assigned to that role. Specifically, we combine the spatial coordinates of all players in role $r$ to represent the characteristic movement patterns of that role. 
We then construct a probability distribution $P_r$ over the spatial grid by calculating the relative frequency of visits to each grid cell. 


We then calculate the Shannon entropy $H(P_r)$ for each role distribution. $E_s$ is defined as:

\begin{equation}
E_s = 1 - \frac{|H(P_1) - H(P_2)|}{\max(H(P_1), H(P_2))},
\end{equation}
where $H(P_1)$ and $H(P_2)$ are the Shannon entropies of the movement distributions for role types 1 and 2, respectively. The spatial overlap $O$ is calculated using Jaccard similarity between the sets of grid cells visited by each role:

\begin{equation}
O = \frac{|C_1 \cap C_2|}{|C_1 \cup C_2|},
\end{equation}
where $C_1$ and $C_2$ are the sets of grid cells visited by players in roles 1 and 2. 

In our task context, high SMS values represent an effective ``divide and conquer'' strategy where both roles maintain similarly thorough coverage patterns (high $E_s$) while minimizing territorial overlap (low $O$). This balance enhances team coordination by ensuring (1) balanced workload distribution across roles through similar exploration complexity and (2) efficient division of the search area.

We note that SED and SMS share a common foundation in information theory, but they serve different analytical objectives: SED focuses on player-to-player comparisons to measure exploration diversity, while SMS examines role-to-role relationships to quantify effective coordination. We also leave the investigation of potential interactions between the two components of SMS to future work.

\paragraph{\textbf{Spatial Proximity Adaptation (SPA)}} SPA quantifies how teams adjust their spatial coordination strategy when game conditions change at the critical threshold. 
Specifically, this metric measures changes in physical proximity between different roles across two distinct phases of the mission. We divide the game into two halves, before and after the critical threshold when red victims are no longer rescuable, and compare the average distance between players of different roles in each phase. SPA is defined as follows:

\begin{equation}
\text{SPA} = \frac{|D_2 - D_1|}{\max(D_1, D_2)},
\end{equation}
where $D_1$ and $D_2$ are the average distances between members of different roles in the first and second halves of the mission, respectively. For each time step, we calculate the distance between every pair of members from different roles, then average these distances across all time steps within each half of the mission.
This normalized difference quantifies the magnitude of change in spatial coordination across mission phases. Higher values indicate greater changes in inter-role proximity between the two halves of the mission, while lower values indicate more stable spatial relationships throughout the task.

\subsubsection{Collective Intelligence (CI)}

CI is typically measured through aggregate values of metrics derived from collaborative processes and has been shown to predict future team behavior and performance across a range of tasks and settings~\cite{riedl2021quantifying,gupta2019digitally}, including search and rescue operations~\cite{eadeh2022good,zhao_teaching_2023}. In this analysis, CI is calculated using three components: effort, defined as the player's travel distance relative to the maximum possible area; skill, measured by time allocated to role-specific actions; and task strategy, evaluated by the proportion of task completions relative to the maximum possible tasks for each role.



\subsubsection{Team Performance} 

Team performance was computed as a weighted performance score based on the successful rescue of different victim types, with weights reflecting the relative priority of each victim category as defined in the task instructions. Specifically, each red victim rescue contributed 60 points, each yellow victim rescue contributed 30 points, and each green victim rescue contributed 10 points to the total team score.

%% file: 4.results.tex
We examine how implicit coordination patterns influence team performance, explore collective intelligence as a mediating mechanism, and analyze their nonlinear and temporal dynamics. This approach reveals not only which coordination dimensions matter most, but also how and when they contribute to effective team performance.

\subsection{Effects of implicit spatial coordination on team performance}

\begin{figure*}[!htbp]
\centering
 \includegraphics[width=1\textwidth]{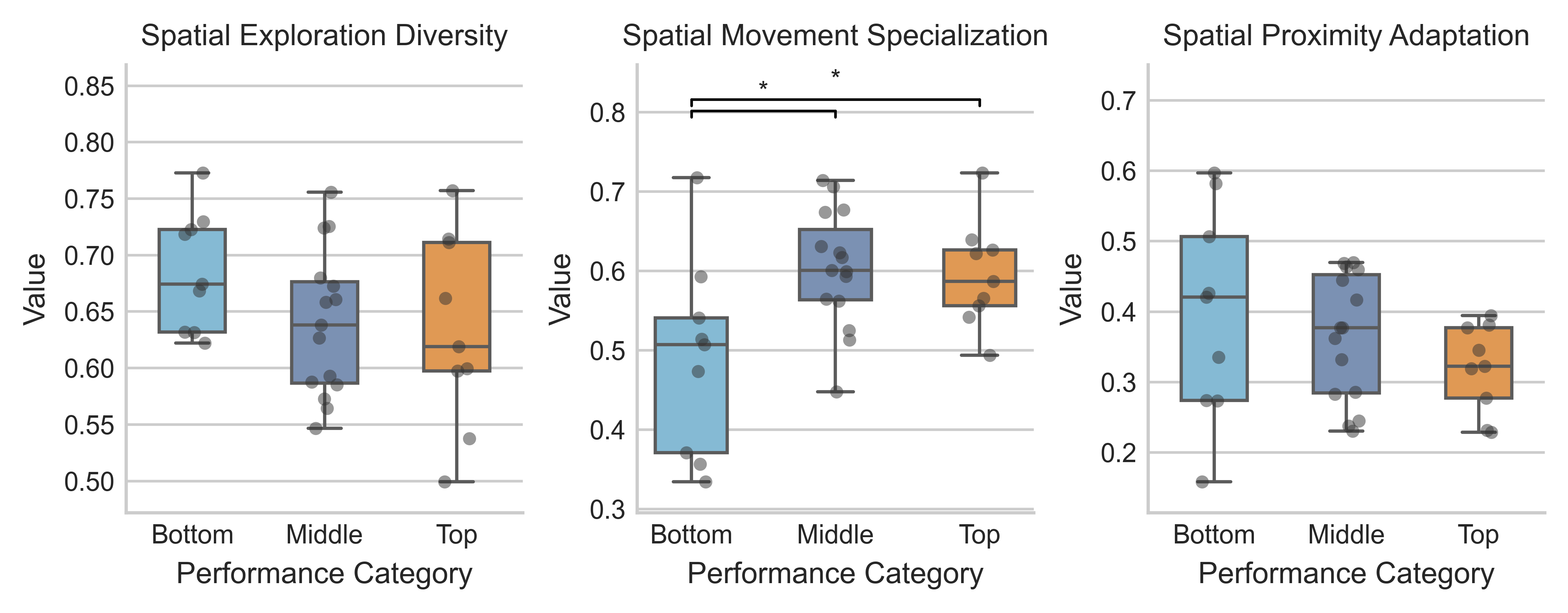}
\caption{Comparison of the spatial coordination metrics across performance groups.}
\label{fig:performance_groups}
\end{figure*}

To answer our first research question regarding how different dimensions of coordination contribute to team performance, we examined the relationships between our three coordination metrics, collective intelligence, and team performance. Table \ref{tab:correlations} presents the correlation matrix for these variables. The analysis showed that spatial movement specialization was significantly positively correlated with team performance ($r = .38$, $p < .05$), indicating that teams with better coordination between the specialized roles achieved higher performance outcomes. By contrast, neither spatial exploration diversity nor spatial proximity adaptation showed significant correlations with performance. We also found that the effective coordination among roles was significantly correlated with collective intelligence ($r = .46$, $p < .01$), which in turn strongly correlated with team performance ($r = .46$, $p < .01$). 

\begin{table}[!htbp]
\centering
\caption{Spearman correlations among coordination metrics, collective intelligence, and team performance (N=34 teams).}
\label{tab:correlations}
\small
\begin{tabular}{@{}lccccc@{}}
\toprule
Variables & 1 & 2 & 3 & 4 & 5 \\
\midrule
1. Spatial Exploration Diversity & 1 & $-0.06$ & $0.29$ & $-0.27$ & $-0.18$ \\
2. Spatial Movement Specialization &  & 1 & $0.08$ & $0.46^{**}$ & $0.38^{*}$ \\
3. Spatial Proximity Adaptation &  &  & 1 & $-0.24$ & $-0.12$ \\
4. Collective Intelligence &  &  &  & 1 & $0.46^{**}$ \\
5. Team Performance &  &  &  &  & 1 \\
\bottomrule
\multicolumn{6}{l}{$^*p < .05$, $^{**}p < .01$, $^{***}p < .001$} \\
 \\
\end{tabular}
\end{table}

To further examine the relationships identified in our correlation analysis, we performed multiple regression with the three coordination metrics as simultaneous predictors (Table \ref{tab:regression_results}). Spatial specialization significantly predicted both collective intelligence ($\beta = 0.05$, $p < .01$) and team performance ($\beta = 108.84$, $p < .05$) when controlling for other coordination dimensions, while neither spatial exploration diversity nor spatial proximity adaptation showed significant effects. The models explained 34\% of the variance in collective intelligence ($F = 5.01$, $p < .01$) and 20\% ($F = 2.45$, $p = .08$) of team performance, suggesting that effective spatial coordination between specialized roles is a critical factor both in enhancing collective intelligence and in influencing team outcomes.

\begin{table}[!htbp]
\centering
\caption{OLS regression results predicting collective intelligence and team performance from coordination metrics.}
\label{tab:regression_results}
\small
\begin{tabular}{@{}lcc@{}}
\toprule
Dependent variable & \begin{tabular}[c]{@{}c@{}}Collective\\Intelligence\end{tabular} & \begin{tabular}[c]{@{}c@{}}Team\\Performance\end{tabular} \\
\midrule
Spatial Exploration Diversity & $-0.02$ & $-19.51$ \\
Spatial Movement Specialization & $0.05^{**}$ & $108.84^{*}$ \\
Spatial Proximity Adaptation & $-0.01$ & $-32.25$ \\
\midrule
$R^2$ & 0.34 & 0.20 \\
F & $5.01^{**}$ & $2.45$ \\
\bottomrule
\multicolumn{3}{@{}l@{}}{\footnotesize $^*p < 0.05$; $^{**}p < 0.01$.} \\
\end{tabular}
\end{table}

Next, we examined how coordination metrics differed across performance groups. Figure \ref{fig:performance_groups} illustrates coordination metrics across teams grouped by team score level (bottom 25\%, middle 50\%, and top 25\%). Mann-Whitney U tests revealed that role spatial coordination was significantly higher in middle and top-performing teams compared to bottom performers ($p < .05$), while spatial exploration diversity and spatial coordination adaptation showed no significant differences across groups. This finding corroborates our correlation and regression results, confirming that role spatial coordination is the primary coordination dimension predicting both collective intelligence and team performance.

\subsection{Collective intelligence as mediator of coordination effects}

To address our second research question regarding how coordination patterns influence performance, we conducted bootstrapped mediation analyses~\cite{hayes2009beyond} to test whether collective intelligence mediates the relationship between coordination metrics and team performance. This method provides more reliable estimates of indirect effects, particularly in smaller samples.

\begin{table}[!htbp]
\centering
\caption{Bootstrapped mediation analysis of coordination metrics on team performance through collective intelligence}
\label{tab:mediation}
\small
\begin{tabular}{lccc}
\toprule
\textbf{Paths} & \textbf{SED} & \textbf{SMS} & \textbf{SPA} \\
\midrule
IV → CI (a) & -0.39 & 0.48** & -0.19 \\
CI → Perf (b) & 1446.18** & 1119.82* & 1409.98** \\
Total effect (c) & -523.97 & 1135.67* & -389.77 \\
Direct effect (c') & 38.77 & 595.03 & -117.59 \\
\midrule
Indirect (a×b) & -562.74 & 540.64 & -272.18 \\
95\% CI & [-1229.83, & [55.19, & [-720.35, \\
& 137.11] & 1268.06] & 166.69] \\
\% mediated & --- & 47.6\% & --- \\
\bottomrule
\end{tabular}

\centering
\vspace{0.2cm}
\footnotesize SED = Spatial Exploration Diversity, SMS = Spatial Movement Specialization, \\
SPA = Spatial Proximity Adaptation, CI = Collective Intelligence, 
Perf = Team Performance. \\ Bootstrapping based on 5000 resamples. *p < .05; **p < .01.
\end{table}

\begin{figure}[!htbp]
\centering
\includegraphics[width=0.95\columnwidth]{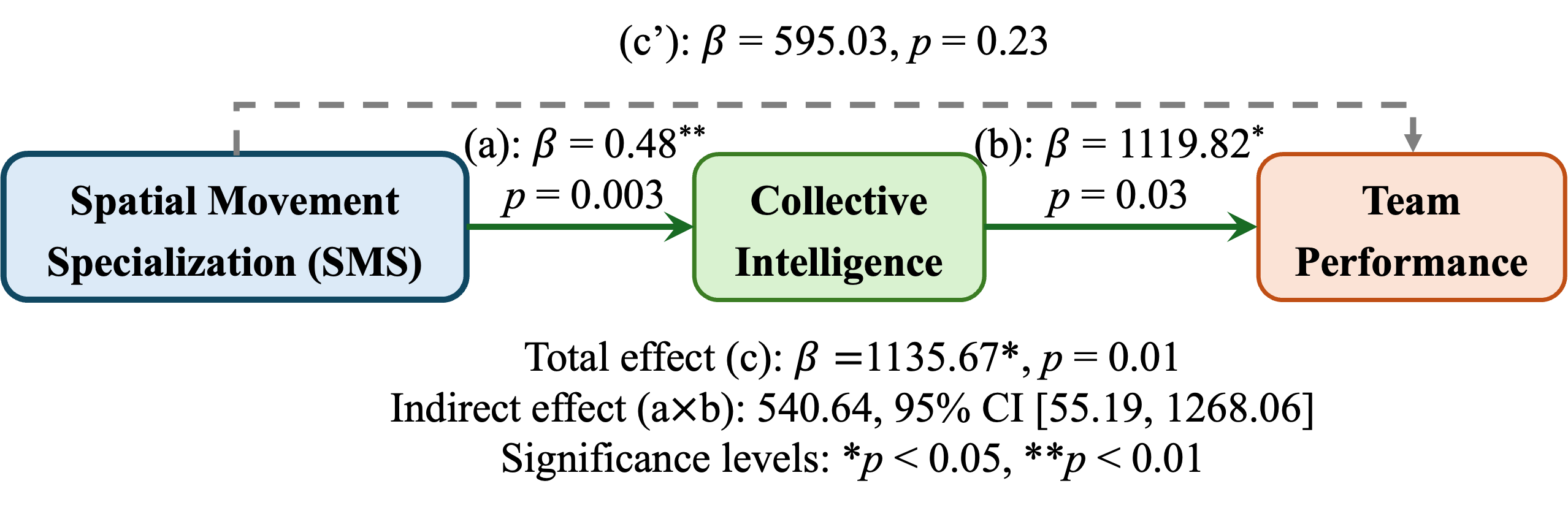}
\caption{Mediation model showing how collective intelligence mediates the relationship between spatial movement specialization and team performance. Path coefficients and significance levels are displayed, with the direct effect (c') becoming non-significant when accounting for the mediator.}
\label{fig:ci-mediation}
\end{figure}

Our mediation analysis revealed that collective intelligence serves as a significant mediator in the relationship between spatial movement specialization and team performance. As shown in Table \ref{tab:mediation} and illustrated in Figure \ref{fig:ci-mediation},  spatial specialization significantly predicted collective intelligence (path a: $\beta = 0.48$, $p < .01$), which in turn predicted team performance (path b: $\beta = 1119.82$, $p < .05$). The significant indirect effect (540.64, 95\% CI [55.19, 1268.06]) accounts for 47.6\% of the total effect. While the total effect of spatial movement specialization on performance was significant ($\beta = 1135.67$, $p < .05$), the direct effect became non-significant when accounting for collective intelligence ($\beta = 595.03$, $p = .23$). The confidence interval for the indirect effect does not contain zero, providing statistical evidence for mediation based on bootstrapping procedures with 5000 resamples. In contrast, we found no significant mediation effects for spatial exploration diversity or spatial proximity adaptation, as their bootstrap confidence intervals included zero. These results suggest that effective coordination between specialized roles enhances team performance primarily by fostering collective intelligence.

\subsection{Nonlinear effects of spatial coordination and temporal dynamics}
Our third research question investigates nonlinear dynamics of coordination patterns, examining whether an optimal level of adaptive spatial proximity exists and how these patterns evolve over time in high-versus low-performing teams.

\begin{table}[t]
\centering
\caption{Quadratic regression analysis of coordination metrics on team performance and comparison across performance groups}
\label{tab:quadratic}
\resizebox{0.5\textwidth}{!}{%
\begin{tabular}{@{}lccc@{}}
\toprule
 & \textbf{SED} & \textbf{SMS} & \textbf{SPA} \\
\midrule
\multicolumn{4}{@{}l@{}}{\textit{Quadratic regression analysis}} \\
\hspace{0.3cm}Constant & 5270.05 & -1332.25 & -326.04 \\
 & (0.172) & (0.192) & (0.501) \\
\hspace{0.3cm}Linear term & -14672.21 & 5354.67 & 4660.33$^{\dagger}$ \\
 & (0.221) & (0.162) & (0.08) \\
\hspace{0.3cm}Quadratic term & 10955.20 & -3906.89 & -6693.82$^{*}$ \\
 & (0.236) & (0.264) & (0.05) \\
\hspace{0.3cm}$R^2$ & 0.06 & 0.21* & 0.13 \\
\hspace{0.3cm}$F$-statistic & 1.03 & 4.01* & 2.31 \\
 & (0.370) & (0.029) & (0.116) \\
\hspace{0.3cm}Optimal value & --- & --- & 0.348 \\
\hspace{0.3cm}Pattern & Non-significant & \begin{tabular}[c]{@{}c@{}}Positive with\\diminishing returns\end{tabular} & \begin{tabular}[c]{@{}c@{}}Inverted\\U-shape\end{tabular} \\
\midrule
\multicolumn{4}{@{}l@{}}{\textit{Performance by group}} \\
\hspace{0.3cm}Low & 519.09 (215.61) & 232.73 (223.07) & 378.18 (270.29) \\
\hspace{0.3cm}Medium & 295.45 (263.07) & 500.00 (237.07) & 567.27 (299.14) \\
\hspace{0.3cm}High & 424.55 (293.20) & 506.36 (264.13) & 293.64 (153.84) \\
\hspace{0.3cm}ANOVA $F$ & 2.06 & 4.58* & 3.48* \\
 & (0.145) & (0.018) & (0.044) \\
\bottomrule
\multicolumn{4}{@{}l@{}}{\footnotesize SED = Spatial Exploration Diversity, SMS = Spatial Movement Specialization, SPA = Spatial Proximity Adaptation.} \\
\multicolumn{4}{@{}l@{}}{\footnotesize Values in parentheses represent p-values for coefficients and standard deviations for group means.} \\
\multicolumn{4}{@{}l@{}}{\footnotesize $^{\dagger}p < .10$; $^{*}p < .05$.} \\
\end{tabular}
}
\end{table}

\begin{figure}[!htbp]
\centering
 \includegraphics[width=0.9\linewidth]{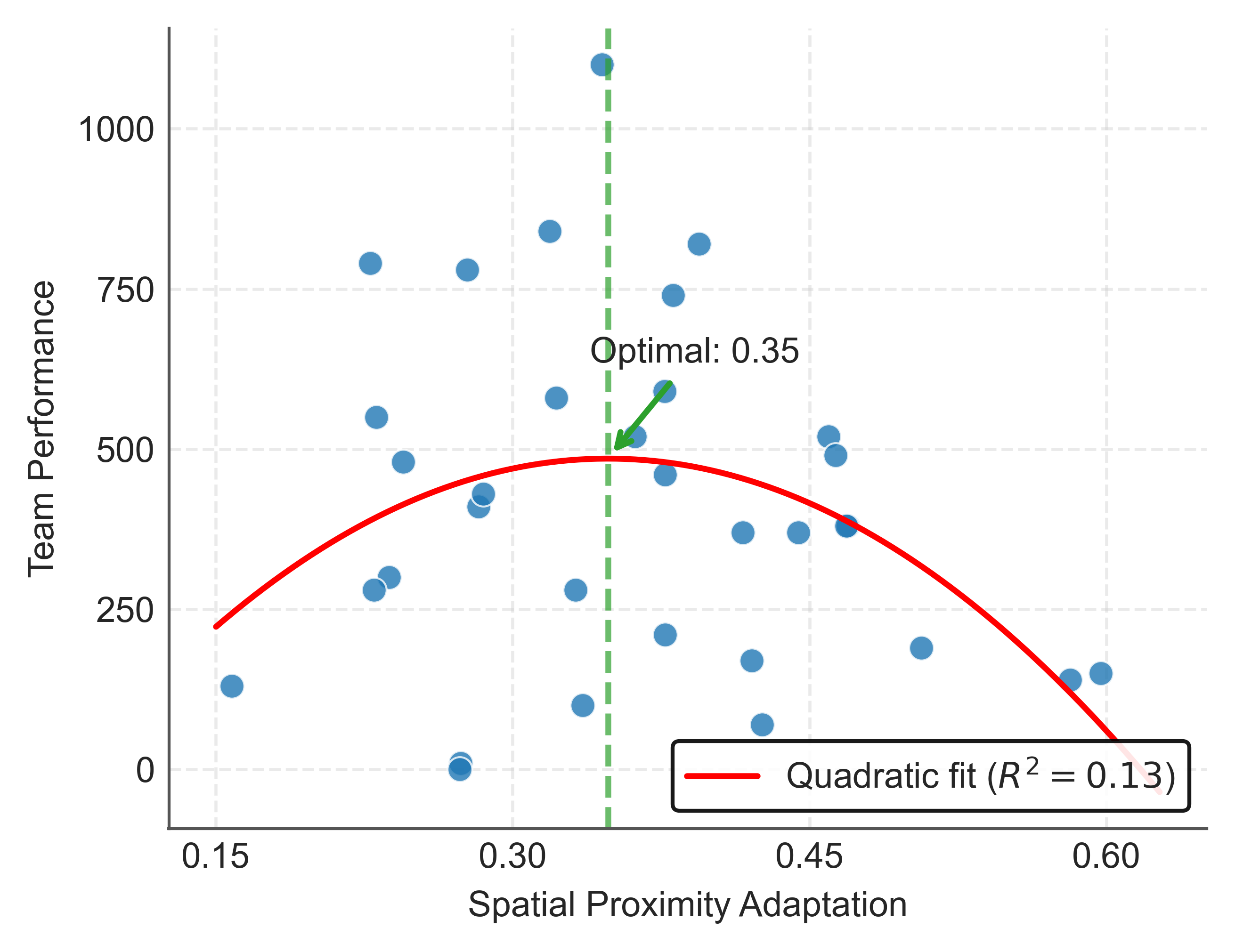}
\caption{Quadratic relationship between spatial proximity adaptation and team performance.}
\label{fig:quadratic-SCA}
\end{figure}

\subsubsection{Nonlinear relationship analysis}
To examine potential nonlinear effects, we conducted quadratic regression analyses with our three metrics as predictors of team performance (Table \ref{tab:quadratic}). The results revealed a marginally significant inverted U-shaped relationship for spatial proximity adaptation ($R^2 = .134$, $F(2,31) = 2.315$, $p = .116$), with the quadratic term approaching conventional significance ($\beta = -6693.82$, $p = .06$). The model included a positive linear coefficient ($\beta = 4660.33$, $p = .085$) and a negative quadratic coefficient, indicating that performance increases with adaptation up to a certain point before declining. As shown in Figure~\ref{fig:quadratic-SCA}, the optimal value of spatial proximity adaptation was calculated at $0.348$ (within the observed range of $0.158$ to $0.597$). This inverted U-shaped relationship was further supported by ANOVA results showing significant performance differences across proximity adaptation groups ($F(2,30) = 3.48$, $p = .044$), with medium adaptation teams ($M = 567.27$) outperforming both low ($M = 378.18$) and high adaptation teams ($M = 293.64$). 

No significant nonlinear relationship was observed for spatial exploration diversity ($R^2 = .06$, $p = .37$). For spatial specialization, despite a significant quadratic model ($R^2 = .21$, $p = .03$), the pattern revealed a positive relationship with diminishing returns, as teams with medium and high spatial movement specialization performed similarly well ($M=500.00$ and $M=506.36$, respectively) and significantly outperformed teams with low specialization ($M=232.73$, $F(2,30) = 4.58$, $p = .02$).

\subsubsection{Temporal patterns of coordination}

\begin{figure*}[!htbp]
\centering
\includegraphics[width=1\linewidth]{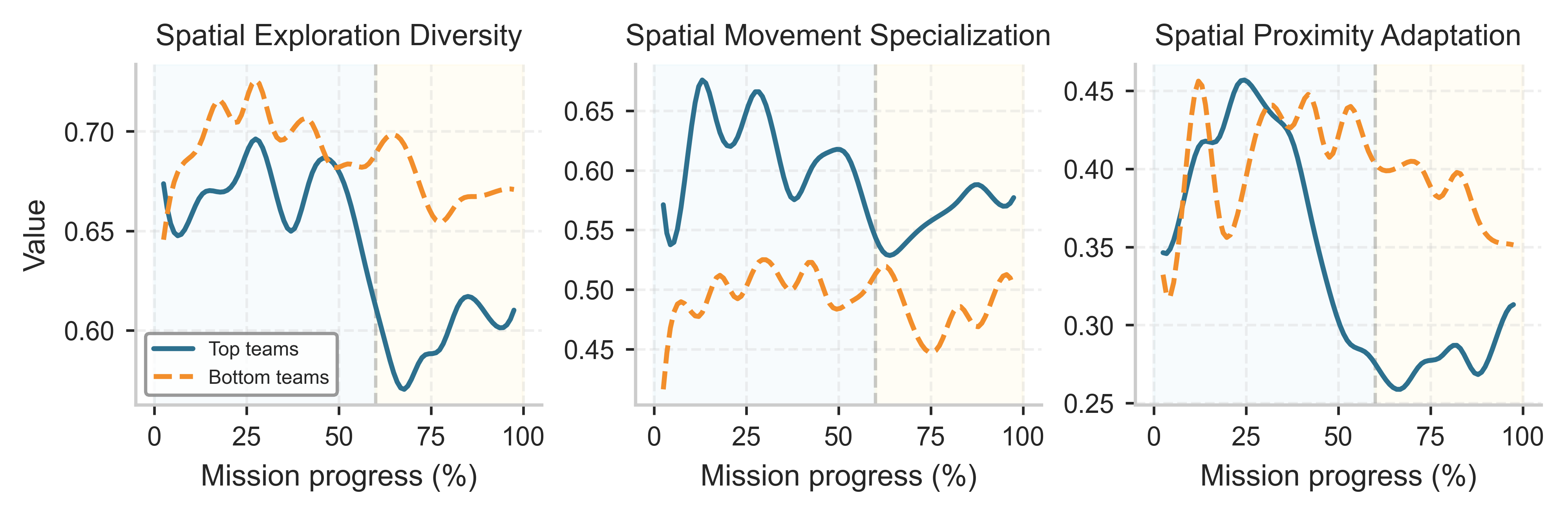}
\caption{Smoothed performance trends of top- and bottom-performing teams over time, represented as a percentage of mission progress across three metrics. Blue and orange shading highlight game dynamics before and after the 3-minute mark of the 5-minute mission (60\% mission), when red victims become non-rescuable.}
\label{fig:metric-over-time}
\end{figure*}

Figure \ref{fig:metric-over-time} illustrates how coordination metrics evolve over time for top and bottom-performing teams. For spatial specialization (middle panel), top-performing teams maintained consistently higher levels throughout the mission compared to bottom-performing teams. Notably, top teams started with higher specialization values (approximately 0.55) that quickly increased to peak levels (around 0.67) during the early phase, while bottom teams began at lower levels (approximately 0.45) and never reached comparable specialization values. This pattern highlights the importance of establishing effective spatial specialization among roles early in the mission.

Spatial proximity adaptation (right panel) reveals an interesting divergence after the critical 60\% mission progress mark. Top-performing teams maintained closer proximity between specialized roles during the first half of the mission (adaptation values peaking around 0.46) when rescuing red victims required the medics and engineers to work adjacent to each other. After the 3-minute mark, when red victims were no longer available according to mission rules, top teams strategically increased the distance between roles (adaptation dropping to about 0.26), allowing them to spread out and efficiently rescue the remaining victims independently. In contrast, bottom-performing teams fail to adapt their spatial configuration to the changing task requirements. This suggests that effective teams dynamically adjust their coordination strategy based on the specific demands of different mission phases. 

For spatial exploration diversity (left panel), both team types showed similar values early in the mission, but a notable divergence occurred around the 50\% mark. Top teams reduced their exploration diversity in the latter half (dropping from approximately 0.68 to 0.60), while bottom teams maintained higher diversity levels. This pattern, combined with the adaptation findings, suggests that successful teams may transition from exploration (searching broadly) to exploitation (focusing on effective strategies) after the mission midpoint, whereas less successful teams continue searching for new approaches throughout the mission without strategically adapting to changing task requirements.
These temporal patterns corroborate the inverted U-shaped relationship between adaptation and performance, showing that optimal coordination requires both appropriate adaptation levels and strategic transitions from exploration to exploitation as the task progresses.

%% file: 5.discussion.tex
In this study, we investigated how implicit spatial coordination influences team performance, with collective intelligence (CI) as a mediating mechanism. Our findings show the significant impact of spatial movement specialization (SMS), which emerged as the only spatial metric that significantly predicts both CI and team performance. In contrast, spatial exploration diversity (SED) and spatial proximity adaptation (SPA) did not exhibit significant linear relationships with these outcomes. We also found that CI partially mediated the relationship between SMS and team performance, accounting for 47.6\% of the total effect. This finding extends previous research on team processes and outcomes by demonstrating that CI serves as a mechanism through which effective role spatial specialization indirectly enhances performance. Our results suggest that when teams coordinate their spatial movements efficiently, they develop stronger CI capabilities, which in turn lead to better performance outcomes.
Interestingly, SPA exhibited a marginal inverted U-shaped relationship with performance, indicating that performance improved up to a certain level of adaptation, but too much adaptation started to reduce effectiveness. Finally, the temporal dynamics of the three spatial metrics clearly demonstrate the distinction between top- and bottom-performing teams. Taken together, these results underscore the importance of specialized role coordination in spatial navigation tasks and highlight how different coordination dimensions dynamically evolve, especially when explicit communication is restricted.

\subsection{Implications}
Our research has several implications for team coordination research more broadly. Building on prior conceptual work on implicit coordination~\cite{rico2008team}, we propose quantifiable process-level metrics that capture how spatial coordination unfolds, particularly in navigation-based, role-specific tasks where team members must coordinate movements without explicit communication. 
By demonstrating that spatial movement specialization significantly predicts CI, we expand our understanding of observable collaborative process metrics~\cite{riedl2021quantifying,zhao_teaching_2023}. Our results suggest that effective coordination emerges not only from aggregated individual behaviors but also from the strategic positioning of team members relative to others with complementary functional roles. Specifically, when team members with complementary roles maintain appropriate spatial relationships, team effectiveness improves. This finding extends transactive memory theory~\cite{wegner1987transactive} beyond knowledge distribution to include physical positioning—teams must coordinate not only what they know but where they position themselves relative to teammates with complementary abilities.

Our findings also provide two complementary insights for designing AI-powered team support systems. First, the mediation of spatial movement specialization (SMS) through CI highlights a beneficial signal for real-time team monitoring. Tracking how team members position themselves in relation to each other to accomplish a shared task, as in SMS, can inform diagnostic AI systems that detect emergent coordination breakdowns and predict team effectiveness, especially in contexts where communication is limited but spatial coordination is critical. Second, the observed inverted U-shaped relationship between spatial proximity adaptation (SPA) and team performance supports prior research suggesting that optimal spatial distance balances stability and flexibility~\cite{gorman2010team, gorman2017understanding}. Our results extend this insight by quantifying spatial role coordination in a 2D navigation task without explicit communication, demonstrating that both under- and over-proximity adaptation are associated with reduced performance. This aligns with findings in coordination dynamics, where high-performing teams maintain stable patterns while selectively adapting to changing task demands. Together, these insights deepen our understanding of adaptive team mechanisms and offer actionable metrics for designing training interventions and intelligent support systems~\cite{gorman2010training}. 

In the context of computational modeling, our findings have implications for the development of machine theory of mind~\cite{gupta2023fostering,nguyen2022theory,nguyen2020cognitive}. The spatial coordination patterns captured by our metrics represent structured, observable indicators of role-based anticipation and behavioral adaptation in human teams. These patterns might serve as useful inputs for AI systems aiming to infer latent cognitive states such as intent, role expectations, or coordination strategies. Incorporating such spatial signals into computational models could support the development of AI agents capable of reasoning not only about individual behaviors but also about the evolving interdependencies and coordination demands in multi-agent and human–AI teaming contexts~\cite{gupta2023fostering,nguyen2024predicting}.

\subsection{Limitations and Future Work}
Our study focused on a specific 2D search and rescue task involving predefined role configurations in teams without explicit communication. While this controlled setting allowed us to isolate spatial coordination mechanisms, caution should be exercised when extrapolating our findings to other team structures or task environments, particularly those involving dynamic roles, explicit communication channels, or different spatial demands.

Second, our analysis is based on cross-sectional data collected from a single task session, which allowed us to examine relationships between coordination, collective intelligence, and performance at a specific point in time. However, this approach limits our ability to observe how these relationships evolve or change over time, which is important for understanding teams as complex, adaptive, dynamic systems operating in constantly changing contexts~\cite{marks2001temporally,ilgen1999teams}. Future longitudinal studies that track teams across multiple sessions or phases of training could provide deeper insight into how spatial coordination evolves and whether targeted interventions lead to persistent improvements in team performance.
Future work will also investigate potential interactions between the components of the metrics. 

Finally, our focus on fully implicit coordination represents one end of the coordination spectrum. Future work should examine how spatial coordination metrics interact with varying levels of explicit communication, potentially revealing optimal combinations of verbal and spatial coordination strategies tailored to different task demands and team compositions~\cite{butchibabu2016implicit,entin1999adaptive}.

%% file: 6.appendix.tex
\section{Team Minimap Instructions} 
The following instructions were provided to participants prior to beginning the task. These instructions described the game interface, task objective, roles, and procedures for rescuing different types of victims. 

\begin{figure*}[h!]
    \centering
    \includegraphics[width=1\linewidth]{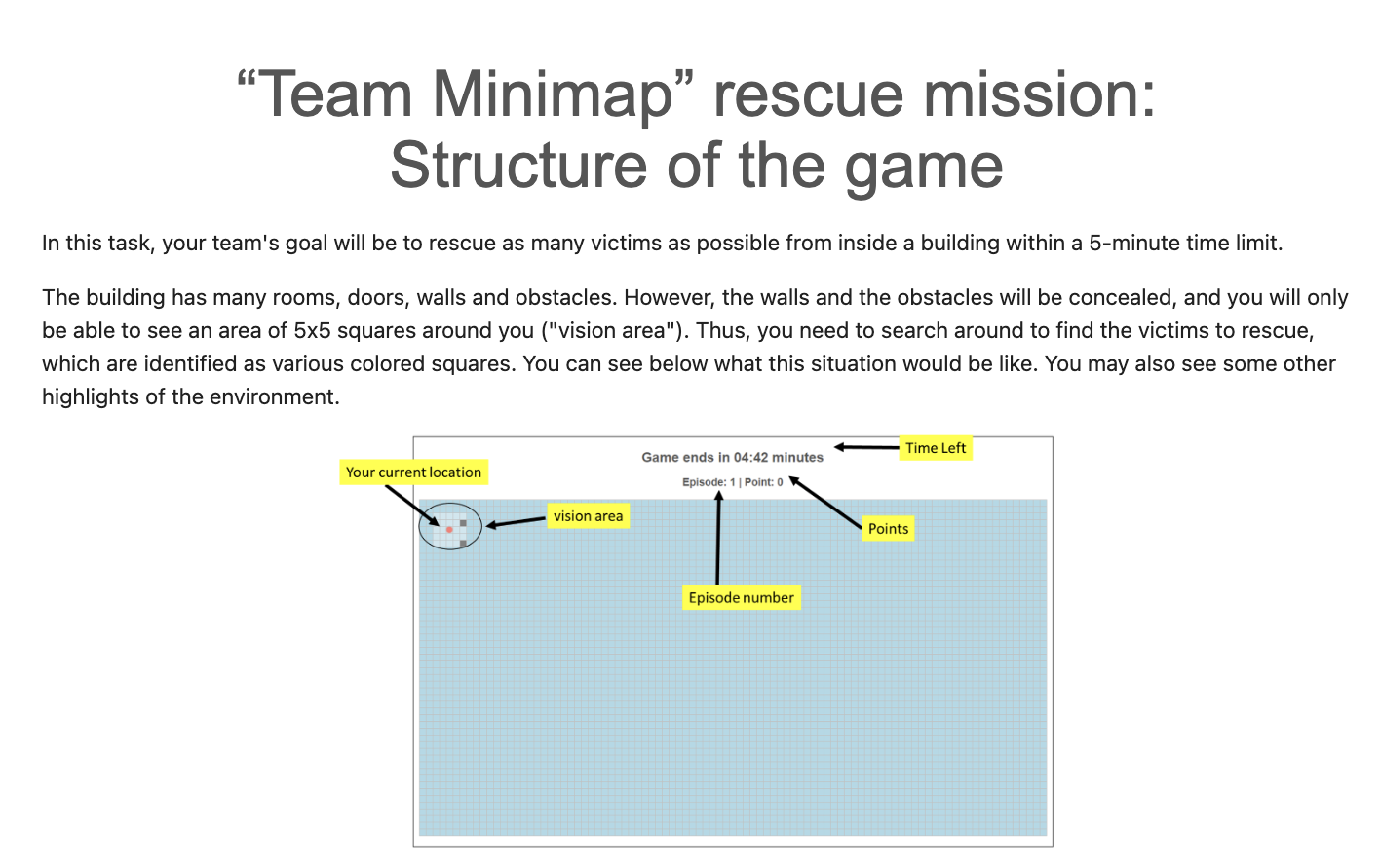}
    \label{fig:appendix-overview}
\end{figure*}

\begin{figure*}[h!]
    \centering
    \includegraphics[width=0.95\linewidth]{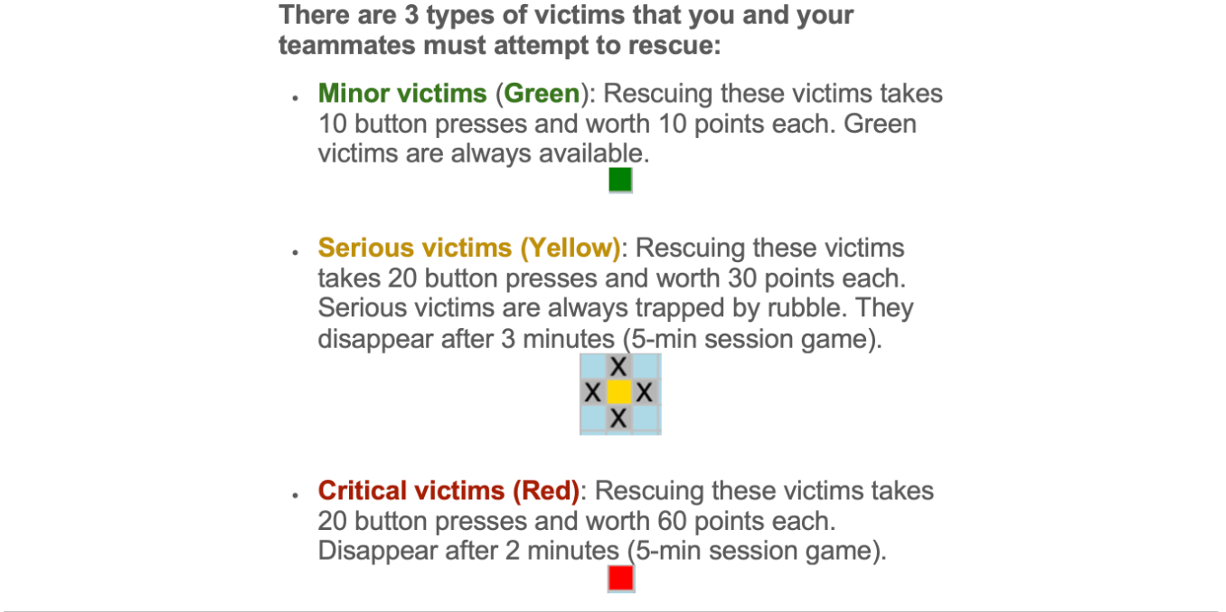}
    \label{fig:appendix-types}
\end{figure*}

\begin{figure*}[h!]
    \centering
    \includegraphics[width=0.95\linewidth]{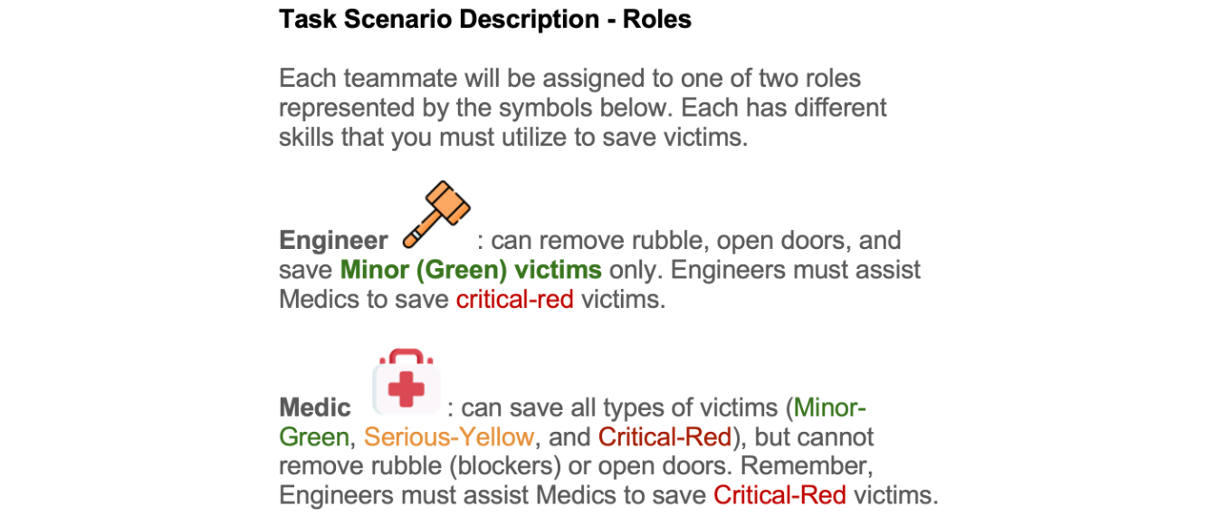}
    \label{fig:appendix-roles}
\end{figure*}

\begin{figure*}[h!]
    \centering
\includegraphics[width=0.95\linewidth]{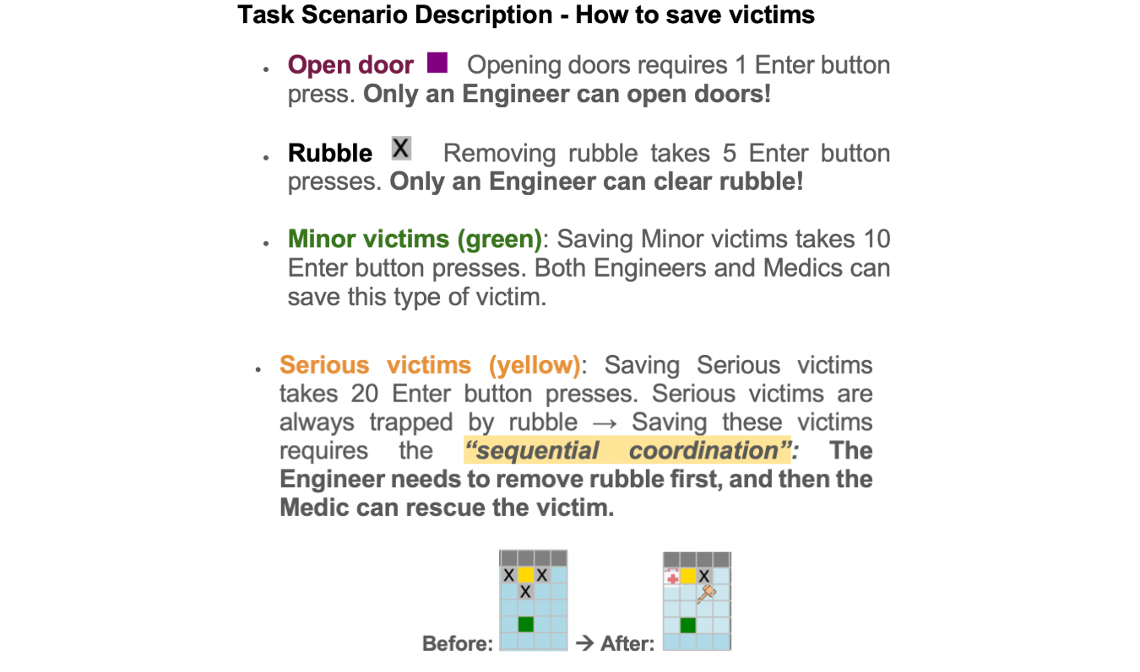}
    \label{fig:appendix-steps}
\end{figure*}

\begin{figure*}[h!]
    \centering
\includegraphics[width=0.95\linewidth]{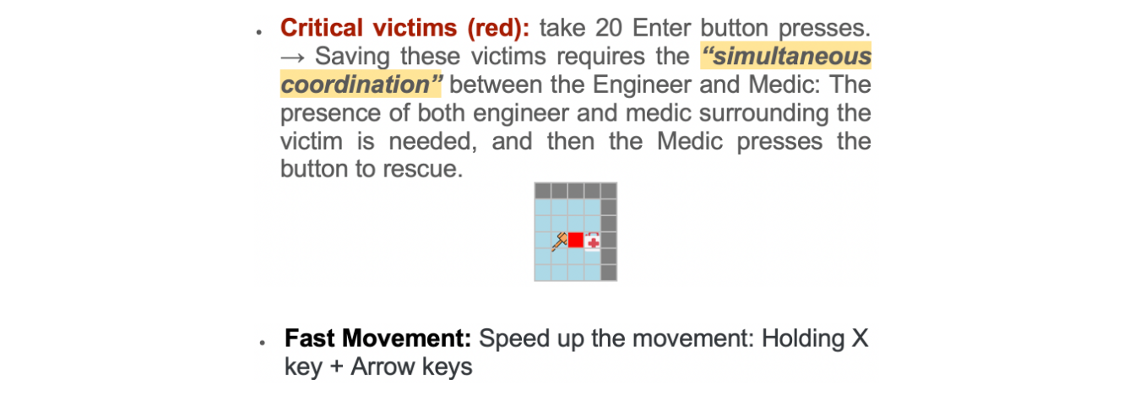}
    \label{fig:appendix-steps}
\end{figure*}